\documentclass{article}
\usepackage{spconf,amsmath,graphicx}

\usepackage{algorithm2e}
\usepackage{pdfpages}
\usepackage{amsfonts}
\usepackage{rotating}
\usepackage{booktabs}
\usepackage{arydshln}
\usepackage{subcaption}
\usepackage{soul}
\usepackage{epstopdf}

\title{Identity and Attribute Preserving Thumbnail Upscaling}

\name{Noam Gat, Sagie Benaim, Lior Wolf}
\address{Tel Aviv University}

\begin{document}

\maketitle
\begin{abstract}
We consider the task of upscaling a low resolution thumbnail image of a person, to a higher resolution image, which preserves the person's identity and other attributes. Since the thumbnail image is of low resolution, many higher resolution versions exist. Previous approaches produce solutions where the person's identity is not preserved, or biased solutions, such as predominantly Caucasian faces. We address the existing ambiguity by first augmenting the feature extractor to better capture facial identity, facial attributes (such as smiling or not) and race, and second, use this feature extractor to generate high-resolution images which are identity preserving as well as conditioned on race and facial attributes. 
Our results indicate an improvement in face similarity recognition and lookalike generation as well as in the ability to generate higher resolution images which preserve an input thumbnail identity and whose race and attributes are maintained. 

\end{abstract}
\vspace{-0.1cm}
\section{Introduction}
\label{sec:intro}

The ability to generate high resolution images from low resolution ones is applicable in many areas such as medicine, astronomy, microscopy and consumer applications. For every  low resolution image, many possible high-resolution versions exists, creating ambiguity.
This is of particular importance when dealing with faces. For instance, a low resolution thumbnail of Barack Obama may be upscaled by existing methods to a face of a Caucasian man~\cite{obamaarticle}.
This lack of fairness could be extremely disturbing.
In a similar way, facial attributes such as smile may not be visible in the low-resolution thumbnail, and one way wish to control this effect. Our method solves this ambiguity by conditioning on the race and facial attributes of the upscaled thumbnail and is also more faithful to the person's identity in the thumbnail. 

The ability to control these properties depends largely on producing informative facial features. Our method augments ArcFace~\cite{deng2019arcface}, a prominent facial feature extractor, with losses that better preserve identity, race and other attributes. 

\section{Related Work}

\noindent\textbf{Super Resolution}\quad
The problem of super-resolution is a  long standing problem in computer vision \cite{farsiu2004advances, park2003super}. In the advent of convolutional neural networks, recent approaches attempt to construct a higher resolution natural image such that its downsampled version matches the low resolution one \cite{dong2014learning, shi2016real, ledig2017photo}. 
Our method is closely related 
to PULSE~\cite{menon2020pulse}. PULSE assumes a prertained StyleGAN generator $G$ and applies projected gradient descent on $G$'s latent space to find a vector $z$ such that $G(z)$ is both natural looking and can be downsampled to match the input low resolution image. PULSE suffers from drawbacks that our framework improves. First, it is biased towards producing Caucasian faces. Second, it produces images which do not preserve identity. 

\noindent\textbf{Facial Feature Extraction}\quad 
Our method incorporates the training of a strong feature extractor, which captures the person's identity.
Typically one uses the representation of a face recognition network that maps the face  into a
feature with low intra-class and large inter-class variance~\cite{parkhi2015deep, sun2014deep, taigman2014deepface, schroff2015facenet}. ArcFace~\cite{deng2019arcface} is a prominent feature extractor that improves the discriminative power
of the model.
Our method improves upon ArcFace on two fronts. First, ArcFace cannot capture the person's race and attributes, which are method improves upon significantly. 
Second, we propose a loss term that better separates the facial features of difference persons while producing similar features for images of the same person. 

\section{Method}
\label{sec:format}

Our method combines image generation with face feature extraction and reasoning. To this end, it builds upon PULSE~\cite{menon2020pulse}. 

\subsection{Super Resolution using PULSE}
\label{sec:pulse}

PULSE receives as input, a low resolution $32 \times 32$ thumbnail image  $I_{LR}$, assumed to be a downscaled version of a higher resolution image $I_{GT}$. It then super-resolves $I_{LR}$ by generating a $1024 \times 1024$ high resolution image $I$ that minimizes: 
\begin{align}
    \mathcal{L}_{DS} = ||DS(I) - I_{LR}||_2 \label{eq:superresolve}
\end{align}
where $DS$ is a bicubic downscaling operation (linear interpolation behaves similarly over different seeds).
To this end, PULSE applies projected gradient descent on the latent space of a pretrained StyleGAN~\cite{karras2019style} generator $G$, to find a latent vector $z$ that satisfies Eq.~\ref{eq:superresolve}, where $I=G(z)$. 

In StyleGAN, $z$ is of dimension $18 \times 512$. Let $z=v_1, \dots, v_{18}$, where each $v_i$ is of dimension $512$. PULSE applies a further `GEOCROSS' regularization which encourages the $v_i$'s to be close to each other on the unit sphere:
\begin{align}
\mathcal{L}_{GS} = \sum_{i<j} \theta(v_i, v_j)^2
\end{align}
where $\theta(a, b)$ denotes the angle between $a$ and $b$. Together with Eq.~\ref{eq:superresolve}, this allows $v_i$'s to vary, thus benefiting from the expressive power of StyelGAN, while not deviating from the natural image manifold of StyleGAN. The overall super-resolution objective, minimized over input $z$, is:
\begin{align}
\mathcal{L}_{SR} = 100\mathcal{L}_{DS} + 0.05 \mathcal{L}_{GS} \label{eq:pulse_opt}
\end{align}

\subsection{Controlling Facial Features}
\label{sec:optimization}

Our method builds upon PULSE by adding a module which accepts StyleGAN's generated face, $I = G(z)$, as input, and enforces additional, user-specified constraints, on $I$. 
An illustration is given in Fig.~\ref{fig:loss}(a). 
$I$ is first passed through a pretrained face alignment module of MTCNN~\cite{zhang2016joint}
and then through a pretrained feature extractor $F$  (described in Sec.~\ref{sec:extractor}), to generate a vector $f_I$ of dimension $512$. 
We define three optimization objectives, which may or may not be optimized, depending on the user's choice. 

The \textbf{feature similarity loss} maximizes the similarity of $I$ with a user specified input image $I_r$. 
Specifically, we define:
\begin{align}
{sim}(i, j, tgt) = \| \sigma \left( \frac{\theta(F(i), F(j)) - \alpha_{thr}}{\gamma_{temp}} \right) - tgt \|
\label{eq:similarity}
\end{align}
to be the similarity measure of images $i$ and $j$ with respect to a target $tgt$, 
where $\theta(F(i),F(j))$ measures the angle between $F(i)$ and $F(j)$ and $\sigma$ is the sigmoid operator. $\alpha_{thr}$ is a threshold angle which is determined experimentally (see appendix \ref{sec:angles}). $tgt$ determines the degree of desired similarity between $i$ and $j$. Specifically, ${sim}(i, j, 0)$ is minimized when $F(i)$ and $F(j)$ are parallel and ${sim}(i, j, 1)$ is minimized when when $F(i)$ and $F(j)$ are orthogonal. 
$\gamma_{temp}$ is a confidence temperature which determines how fast the confidence increases when the distance from $\alpha_{thr}$ increases. 
As we wish $I_r$ and $I$ to be similar, we define the objective:
\begin{align}
\mathcal{L}_{sim}(I_r) = sim(I_r, I, 0)
\end{align}

The \textbf{target race loss} maximizes the probability that $I$ has a user specified race $r$. We use a pretrained race classifier $C_R$ (as in Sec.~\ref{sec:extractor}), which accepts as input, the facial features of an image, and predicts its race. 
Specifically, let $C_R(f_I)[r]$ indicate the probability of attribute $r$, which we wish to be as close to $1$ as possible, and so define:
\begin{align}
\mathcal{L}_{race}(r) = || C_R(f_I)[r] - 1||_2
\end{align}

The \textbf{desired attributes loss} assumes a set of attribute-value pairs AV which indicate desired attributes of the generated image I. A `smiling' attribute, and a value of 1 (resp. 0), indicates a smiling (resp. not smiling) person. We assume a pretrained attribute classifier $C_A$ such that $C_A(f_I)[att]$ indicates the probability that attribute $att$ is present in $I$. $val$ is assumed to be $0$ or $1$ (non-existing or existing). We define:
\begin{align}
\mathcal{L}_{att}(AV) = \frac{1}{|AV|} \sum_{(att, val) \in AV} ||C_A(f_I)[att] - val||_2 \label{eq:att_eq}
\end{align}
$C_A$ is trained as in Sec.\ref{sec:extractor}. 
The overall objective is:
\begin{align}
\min_z[\mathcal{L}_{SR} + \lambda_1 \mathcal{L}_{sim}(I_r) +  \lambda_2 \mathcal{L}_{race}(r) + \lambda_3 \mathcal{L}_{att}(AV)] \label{eq:overall}
\end{align}
$\lambda_1-\lambda_3$ are hyperparameters which may be $0$ when not optimized. 
As in PULSE, we reject $I$ if, at the end of optimization, any of $\mathcal{L}_{sim}, \mathcal{L}_{race}, \mathcal{L}_{att}$ are above a constant ($=0.1$).

\subsection{Face Feature Extractor}
\label{sec:extractor}

A core component of our method is our face feature extractor. We build on ArcFace~\cite{deng2019arcface}, a prominent face feature extractor.

\begin{figure*}[t]
  \centering
\begin{tabular}{cc}
\includegraphics[width=0.95\linewidth]{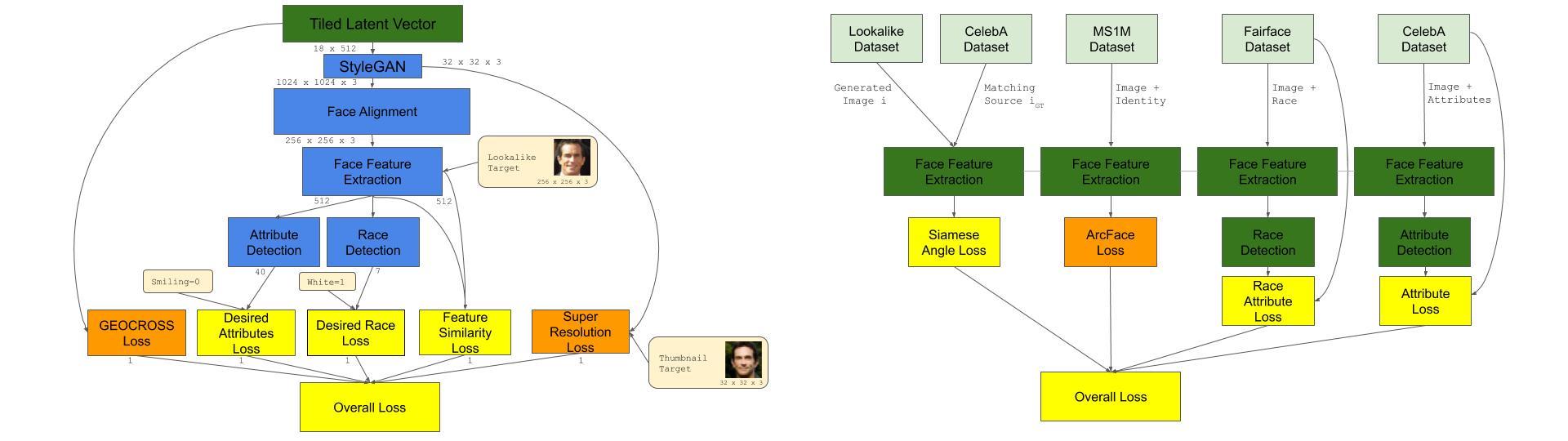} \\
\vspace{-0.3cm}
(a) ~~~~~~~~~~~~~~~~~~~~~~~~~~~~~~~~~~~~~~~~~~~~~~~~~~~~~~~~~~~~~~~~~~~~~~~~~~~ (b) \\
\end{tabular}
\caption{The optimization scheme (a) and training scheme (b) as described in Sec.~\ref{sec:optimization} and  Sec.~\ref{sec:extractor} respectively. In green are optimized components and in blue are pretrained components. In yellow are losses unique to our formulation or the overall loss and in orange are losses used in our framework and in previous work (either PULSE for (a) or ArcFace for (b)). 
For (a), in light yellow are example of user-defined inputs and in small font beside each block are its output dimension.   
For (b), in light green are the datasets used for each branch. Note that the same feature extractor is trained for all four branches.}
\vspace{-0.1cm}
\label{fig:loss}
\vspace{-0.4cm}
\end{figure*}

For image $i$, features $f_i$ are extracted using $F$ as in Sec~\ref{sec:optimization}. 
$F$ is of the same ResNet architecture as in ArcFace~\cite{deng2019arcface}.  $F$ is trained using four objectives, illustrated in Fig.~\ref{fig:loss}(b), and described below. $N$ denotes the batch size.

\noindent\textbf{ArcFace loss}\quad
We consider the feature extraction objective of ArcFace~\cite{deng2019arcface}. Specifically, assume each image belongs to one of $n$ classes. For each image $i$, let $f_i \in R^d$ be its extracted feature as described above. Let $W \in \mathbb{R}^{d \times n}$ be a projection weight matrix and assume $f_i$ and $W$ are L2 normalized. For $j \in [1..n]$, we define $\theta^i_j = \arccos(W^T_j \cdot f_i)$, which represents the angle between the feature $f_i$ and the weight of class $j$. 
Let $y_i$ be the ground truth class of image $i$. The ArcFace loss is: 
\begin{align*}
{\mathcal{L}_{arc}} =-\frac{1}{N}\sum_{i=1}^{N}\log\frac{e^{s(\cos(\theta^i_{y_i}+m))}}{e^{s(\cos(\theta^i_{y_i}+m))}+\sum_{j=1,j\neq  y_i}^{n}e^{s\cos\theta^i_{j}}}
\label{eq:arcface}
\end{align*}
where $m$ is an angular margin penalty and $s$ is a feature scaling operator. The ArcFace objective is trained on the MS1M dataset~\cite{guo2016ms}, with the default train-test splits. 
We now describe additional objectives, unique to our formulation.

\noindent\textbf{Siamese angle loss}\quad 
As shown in Sec.~\ref{sec:results}, using $F$ trained on $\mathcal{L}_{arc}$ only, results in an image $I$ which does not capture $I_{GT}$'s facial features well.
To solve this, we first generate a `lookalike dataset' follows: Each image, $i_{GT}$, in CelebA's training set,  is first downscaled to a $32 \times 32$ resolution image $i_{LR}$. We then apply the optimization of Sec.~\ref{sec:pulse} to generate a higher resolution image $i$ using the current state of $F$.   
Specifically, we optimize for $\mathcal{L}_{SR} + \mathcal{L}_{sim}(i_{LR})$. We also apply the same optimization on a randomly selected image $i'_{GT}$, of the same identity as $i_{GT}$, to generate image $i'$.

As $i_{GT}$ expresses different facial characteristics from $i$, we force $F$ to minimize their similarity. 
The same holds to $i$ and $i'$ which express different facial characteristics.
Lastly, we force $F$ to maximize the similarity between $i$ and $i_T$, a horizontal flip of $i$.
We define $\mathcal{L}_{siam}$ as:
\begin{small}
\begin{align*}    
     \frac{1}{N} \left[ \sum_{i, i_{GT}} {sim}(i_{GT}, i, 1)+\sum_{i, i'} {sim}(i, i', 1)+\sum_{i, i_T} {sim}(i, i_T, 0) \right]
\end{align*}
\end{small}
As $F$ improves, using it to optimize $\mathcal{L}_{SR} + \mathcal{L}_{sim}(i_{LR})$, results in faces whose characterstics are closer to $i_{GT}$. 
As long as $F$ is not too strong, we found experimentally, that augmenting the ‘lookalike dataset' with additional samples generated with such stronger $F$, improves the feature extractor, as it forces it to converge to facial features which are even closer to those of $i_{GT}$. See additional details in the appendix \ref{sec:angles}.

\noindent\textbf{Race and attributes losses}\quad 
Consider a feature extractor $F$ trained to minimized $\mathcal{L}_{arc} + \mathcal{L}_{siam}$ as described above. For the each image $i$ in the FairFace dataset~\cite{karkkainen2019fairface}, we extracted facial features $f_i$ using $F$. We then trained an MLP $M$ on top of the $f_i$'s to classify the correct race using cross-entropy loss. $M$ achieved only $25\%$ accuracy on the test set of FairFace. 

To solve this, 
on the FairFace dataset, we train a classifier $C_R$, to take $f_i$ as input, and predict its ground truth race $r_i$: 
\begin{equation}    
    \mathcal{L}^{class}_{race} = \frac{1}{N} \sum_{i} CE(C_R(F(i)), r_i)
\end{equation}
where $CE$ is the cross entropy loss. $F$ and $C_R$ train jointly and $C_R$ achieves 65\% top-1 accuracy.
As shown qualitatively in Sec.~\ref{sec:results}, this is sufficient to control the race of the generated faces.
In a similar manner, a classifier $C_A$ is trained together with $F$ to correctly classify facial attributes, given features $f_i$. We use CelebA training images and associated attributes.
To correct class imbalance, we additionally apply Inverse Class Frequency~\cite{cui2019class}, based on CelebA's statistics (see more details in the appendix \ref{sec:imbalance}). We found empirically, that training $C_A$ along with $F$ improves $F$'s ability to capture facial attributes. In particular, images generated by optimizing Eq.~\ref{eq:att_eq} as part of Eq.~\ref{eq:overall} more closely capture the desired properties optimized for. $\mathcal{L}^{class}_{attributes}$ is defined as $\mathcal{L}^{class}_{race}$, but where $C_R$ is replaced by $C_A$.
Our final objective, minimized over $F, C_R, C_A$, is: 
\begin{align}
\mathcal{L}_{arc} + 10 \mathcal{L}_{siam} +  10 \mathcal{L}^{class}_{race} + 10 \mathcal{L}^{class}_{attribute}
\label{eq:fulltrain}
\end{align}

\section{Results}
\label{sec:results}

Our results explore the added capabilities of $F$ through the additional losses defined in Sec.~\ref{sec:extractor} as well the effect on the optimization described in Sec.~\ref{sec:optimization}. 

\subsection{Face similarity recognition and lookalike generation}

To generate lookalike faces, $F$ must correctly distinguish between same and different persons.
We denote by $F_{ours}$, a feature extractor
trained as in Sec.~\ref{sec:extractor} using Eq.~\ref{eq:fulltrain} and by $F_{baseline}$,
a feature extractor trained on $\mathcal{L}_{arc}$ only as in ArcFace~\cite{deng2019arcface}. 
As in ArcFace~\cite{deng2019arcface}, two images are considered similar if the angle between their extracted features is small.

We consider the following datasets, each  of $3,000$ image pairs:
\textit{CelebA-Same}: Two randomly chosen images of the same person from CelebA test set.  \textit{CelebA-Diff}: Two randomly chosen images of different persons from CelebA test set.
\textit{PULSE-Pairs}: A randomly chosen image $I_{GT}$ from CelebA test set and a PULSE-generated image $I$ from its thumbnail $I_{LR}$ (using Eq.~\ref{eq:pulse_opt}). 
\textit{Lookalike-Baseline}: As in PULSE-Pairs, but where $I$ is replaced with an image optimized for $\mathcal{L}_{SR} + \mathcal{L}_{sim}(I_{GT})$, 
where 
$F_{baseline}$ is used. 
\textit{Lookalike-Ours}: As in Lookalike-Baseline but where $F_{ours}$ is used. 
For CelebA-Diff, PULSE-Pairs and Lookalike-Baseline, image pairs are considered of different persons and for the rest, they are considered the same.

\begin{table}
\begin{center}
 \begin{tabular}{lccc} 
 \toprule
 & Dataset &$F_{baseline}$ & $F_{ours}$ \\ 
\midrule
& CelebA-Same &51.5$^\circ \pm$13.4 &51.4$^\circ \pm$12.2 \\
& CelebA-Diff &89.8$^\circ \pm$4.4 &88.9$^\circ \pm$5.3 \\
& PULSE-Pairs &81.3$^\circ \pm$5.9 &92.0$^\circ \pm$5.4 \\
& Lookalike-Baseline &76.3$^\circ \pm$6.0 &90.0$^\circ \pm$5.4 \\
& Lookalike-Ours &45.7$^\circ \pm$6.1 &34.8$^\circ \pm$3.8 \\
\midrule
Opt 1 & CelebA-Same+Diff &96.6\% &96.8\% \\
Opt 1 & Lookalike-Baseline &32.4\% &98.9\% \\
Opt 2 & CelebA-Same+Diff &94.5\% &96.6\% \\
Opt 2 & Lookalike-Baseline &88.0\% &99.4\% \\
\toprule
\end{tabular}
\end{center}
\vspace{-0.7cm}
\caption{Top: Reported mean and SD angles for every dataset. 
Bottom: Separation accuracy for the optimal $\alpha_{thr}$. }
\label{tab:pairs_similarity}
\end{table}

For every dataset, we consider the average similarity image pairs ($i_1$, $i_2$) by considering the angle between $F(i_1)$ and $F(i_2)$. A histogram of angles is shown in the appendix \ref{sec:angles}. The mean and standard deviation (SD) is reported in Tab.~\ref{tab:pairs_similarity}. 
Unlike $F_{basline}$, $F_{ours}$ creates similar distributions for CelebA-Diff, PULSE-Pairs and Lookalike-Baseline datasets, with a mean of about $90^{\circ}$, indicating minimum similarity. At the same time, $F_{ours}$ gives similar distribution for CelebA-Same as $F_{basline}$ (mean of $51.4^{\circ}$ to $51.5^{\circ}$ respectively).

Next, we check if a single angle $\alpha_{thr}$ exists for separating the features of CelabA-Same, CelabA-Diff and Lookalike-Baseline datasets. 
In Tab.~\ref{tab:pairs_similarity}, we consider two options for finding $\alpha_{thr}$. Opt 1: using only  CelabA-Same and CelabA-Diff and Opt 2: using all three datasets. A brute-force threshold search is conducted.
For both options, we test on all three datasets. Accuracy is measured as the percentage of images correctly classified as same or different, depending on the dataset. 

For $F_{baseline}$, using Opt 1, an accuracy of 32\% is achieved on Lookalike-Baseline. 
With Opt 2, it is possible to get an accuracy of $88\%$, but the accuracy on the CelebA-Same and CelebA-Diff reduces from $96.5\%$ to $94.5\%$, which is significant. 
In contrast, $F_{ours}$ results in at least $96\%$ accuracy on CelebA-Same and CelebA-Diff and at least $98\%$ accuracy on Lookalike-Baseline using either Opt 1 or Opt 2. 

\begin{figure*}
  \centering
  \vspace{-0.2cm}
  \includegraphics[width=0.85\linewidth]{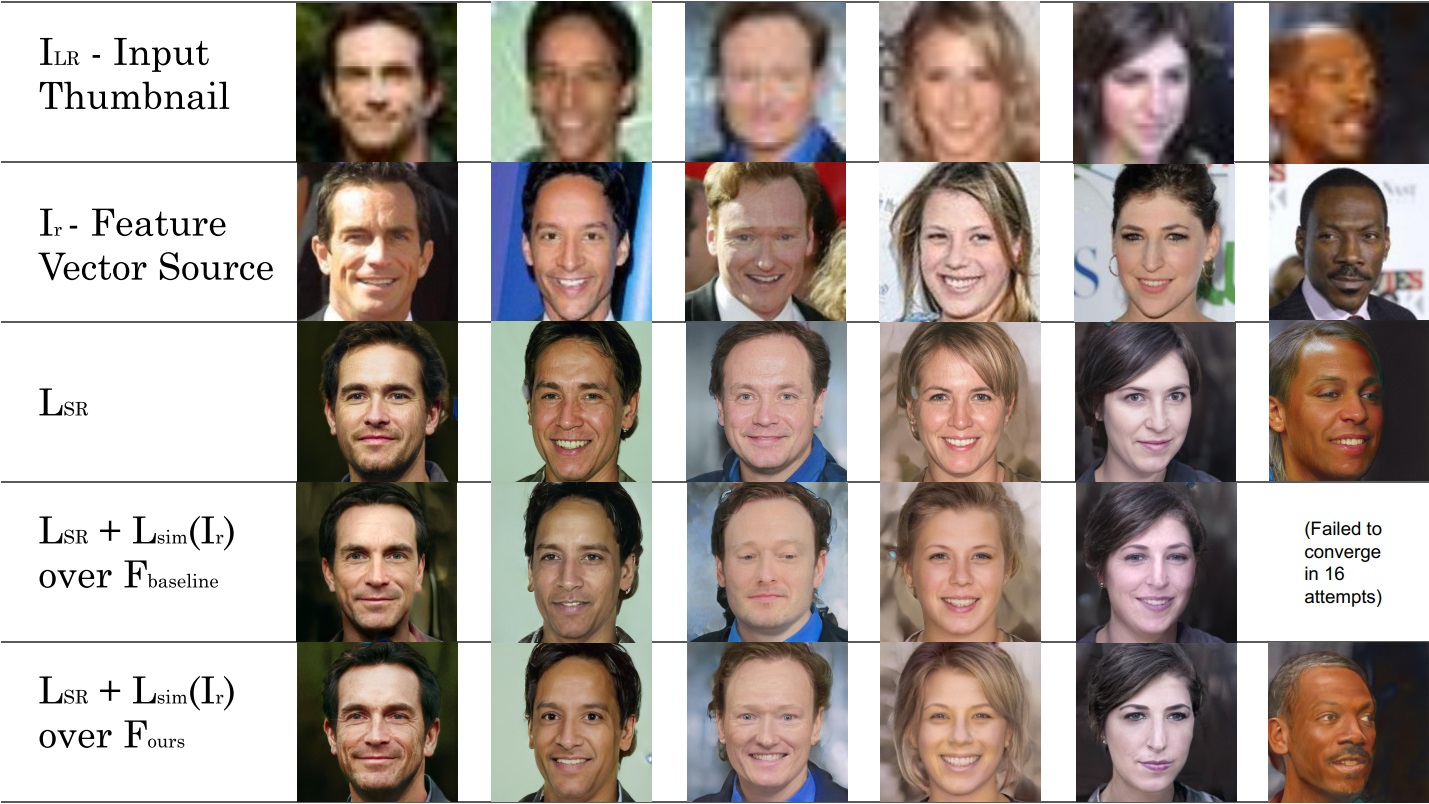}
 \vspace{-0.1cm}
\caption{Comparison of lookalike face generation results. }
\vspace{-0.7cm}
\label{fig:lookalike}
\end{figure*}

In Fig.~\ref{fig:lookalike}, qualitative comparison is shown when generating lookalike images either using $\mathcal{L}_{SR}$ or using $\mathcal{L}_{SR} + \mathcal{L}_{sim}(I_{r})$ with $F_{ours}$ or $F_{baseline}$. Two random images of the same person are taken from CelebA-Same, one is downsampled to a $32 \times 32$ thumbnail $I_{LR}$ and the other acts as $I_r$ for $\mathcal{L}_{sim}$. As can be seen, our method's generated images are much closer to the $I_{LR}$'s facial attributes.

\subsection{Race and attributes re-identification}

To judge the ability to control race, we propose a race re-identification task: given an image with a labelled race, what percentage the generated images are correctly classified as the same race by an off the shelf race classifier? We use 1,000 images from the FairFace testset. 
For each image $I$, let $I_{LR}$ be its lower resolution thumbnail and $r$ its race. 
We consider $I$ generated from $I_{LR}$ using four strategies as shown in Tab.~\ref{tab:race_tab}. 

\begin{table}
\begin{center}
 \begin{tabular}{ccc} 
 \toprule
$F$ & Loss & RE-ID Accuracy \\ \midrule
- & $\mathcal{L}_{SR}$ &20.5\% \\
$F_{baseline}$ & $\mathcal{L}_{SR} + \mathcal{L}_{sim}(I_{LR})$ &40.1\% \\
$F_{ours}$ & $\mathcal{L}_{SR} + \mathcal{L}_{sim}(I_{LR})$ &53.5\% \\
$F_{ours}$ & $\mathcal{L}_{SR} + \mathcal{L}_{race}(r)$ &92.1\% \\
\bottomrule
\end{tabular}
\end{center}
\vspace{-0.4cm}
\caption{Race re-identification accuracy.}
\label{tab:race_tab}
\vspace{-0.4cm}
\end{table}

Our method enables the control of generated images through and explicit race specification through $\mathcal{L}_{race}$ and input face similarity through $\mathcal{L}_{sim}$. As can be seen in Tab.~\ref{tab:race_tab}, both of these losses encourage better identification of race.

\begin{figure*}
  \centering
  \includegraphics[width=0.85\linewidth]{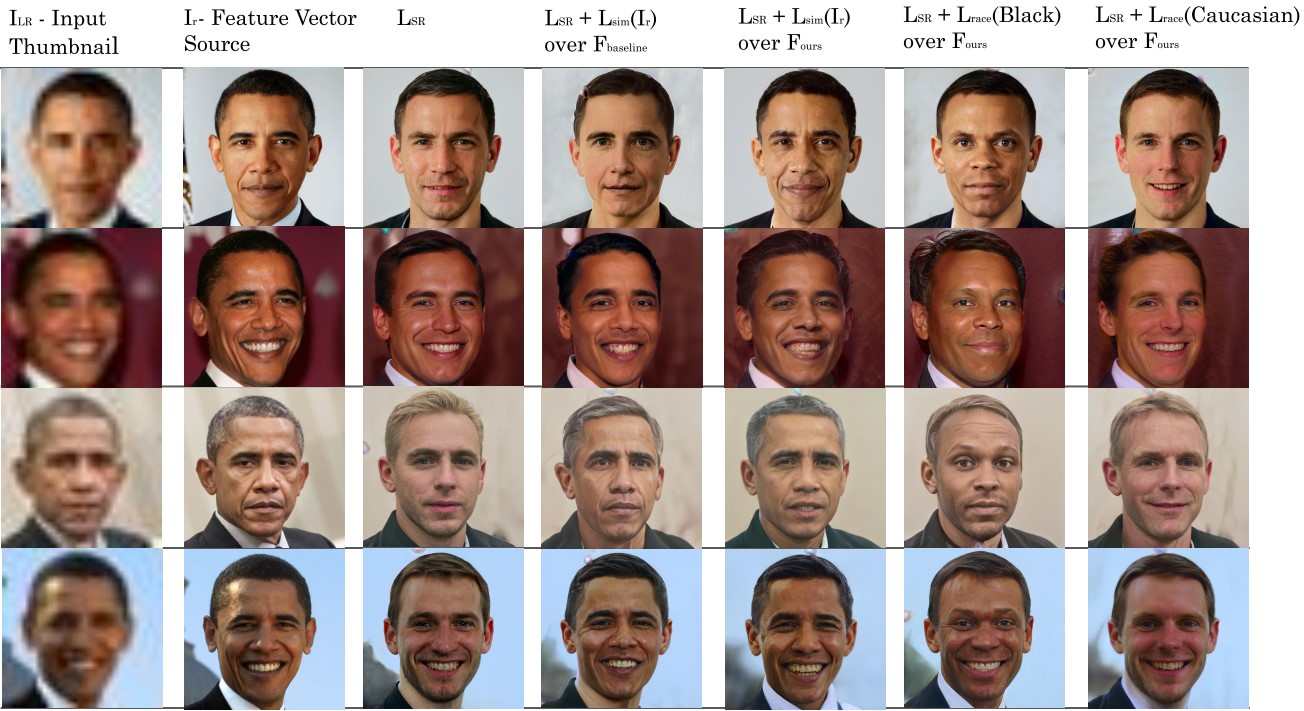}
\caption{Images generating using Barack Obama's thumbnail.}
\vspace{-0.2cm}
\label{fig:qualitative_race}
\end{figure*}

Qualitatively, we revisit one of the criticisms of the PULSE. CelebA is not racially balanced. As StyleGAN is trained on CelebA, generated images are biased accordingly, resulting in Caucasian faces being produced more often. Therefore, many PULSE-generated faces did not match the race of the original image.
By using target race attributes, we can control the race of the generated faces. In Fig.~\ref{fig:qualitative_race}, four faces are generated using Barack Obama's thumbnail, with either a Black or Caucasian race attribute. Our model generates an image with the correct race.

\begin{figure*}
\centering
\includegraphics[width=0.85\linewidth]{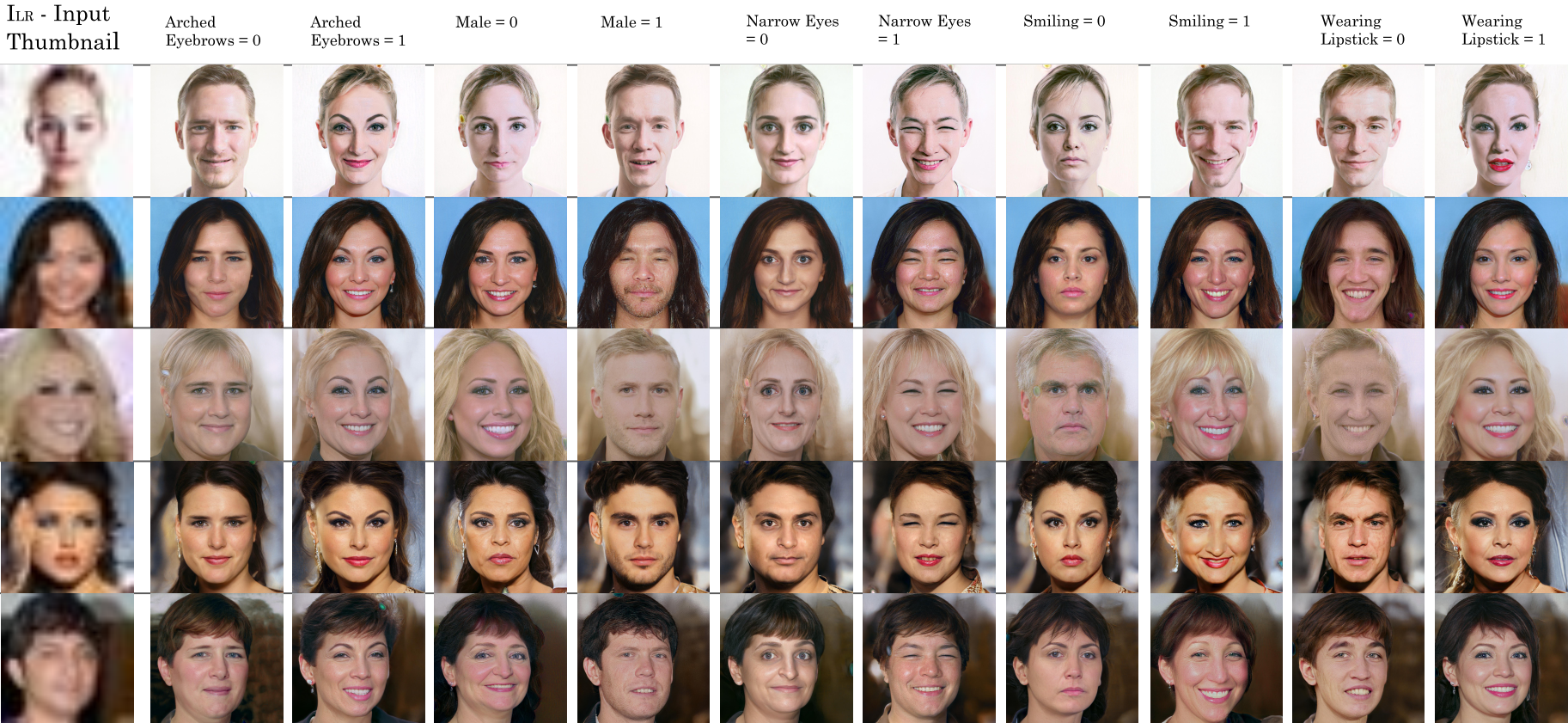}
\caption{Images generated using $\mathcal{L}_{SR} + L_{att}(a)$ for various attribute value pairs}
\label{fig:atts_fig}
\end{figure*}

\subsection{Attribute Re-identification Results}

Our method can also be used to control attributes such as the person's `smile' or gender. 
Similar to the race re-identification task discussed, we test our generated results on five attributes from the CelebA attribute set: Arched Eyebrows, Male, Narrow Eyes, Smiling, Wearing Lipstick. For each attribute, we measure the percentage of correctly generated attributes, based on an external attribute detection algorithm. The average score over the above five attributes can be seen in Tab~\ref{tab:attr_tab_reid} for four different strategies. Using $F_{ours}$ improved the re-identification accuracy from 70.9\% to 79.6\% without adding additional information to the task. Using $\mathcal{L}_{att}$ on a given attribute-value pair $a$ gives a significant increase in accuracy, when averaged over the above mentioned attributes. Qualitative results for different attribute value pairs are shown in Fig.~\ref{fig:atts_fig} for $\mathcal{L}_{SR} + \mathcal{L}_{att}(a)$, when using $F_{ours}$.

\section{Conclusion}

In this work we introduced a method to upscale a low resolution thumbnail while preserving the person's identity and attributes. Through a series of quantitative and qualitative experiments, we showed that our method can maintain the person's race as well as specified attributes, while preserving the image identity. To this end, we showed that a sufficiently strong facial feature extractor is required. This was achieved by modifying the ArcFace features using additional losses, which better preserve identity and other attributes.

\bibliographystyle{IEEEbib}
\bibliography{strings,refs}

\clearpage
\appendix

\section{Training Algorithm}

In Sec.~\ref{sec:extractor}, we note that 
augmenting the ‘lookalike dataset’ with additional samples improves the feature extractor. 
Specifically, Alg-F in Algorithm \ref{alg:alg_f} describes the process of training the feature extractor $F$.

\begin{algorithm}
\SetAlgoLined
\caption{Alg-F}

\textbf{Input:} $F$, augmented-lookalike-dataset \\
\textbf{Output:} $F$\\
 \For{epoch in epochs}{
  $\theta_{thr}$ = calc-best-angle-threshold(CelebA)\;
  \For{b1,b2,b3,b4 in batches}{
    calculate $\mathcal{L}_{arc}$ with b1\;
    calculate $\mathcal{L}^{class}_{race}$ with b2 and $F$\;
    calculate $\mathcal{L}^{class}_{attributes}$ with b3 and $F$\;
    calculate $\mathcal{L}_{siam}$ with b3, $F$, augmented-lookalike-dataset and $\theta_{thr}$\;
    optimize $F$, $C_R$, $C_A$ using Eq.~10
  }
 }
 \label{alg:alg_f}
\end{algorithm}

Each mini-batch of $b_1$-$b_4$ is taken from its corresponding dataset. The input consists of the feature extractor (random at initialization) and an augmented lookalike dataset.
calc-best-angle-threshold stands for running a brute-force threshold search to find the best linear separation threshold angle $\alpha_{thr}$, which achieves the highest classification accuracy on the CelabA-Same and CelabA-Diff datasets together (where train images instead of test images are used). We note that in the first meta-epoch of Algorithm \ref{alg:high_level}, when calling Alg-F, we do not calculate and optimize for $\mathcal{L}_{siam}$ (its value is taken to be 0). 

The overall training procedure is then described in Algorithm \ref{alg:high_level}.
\begin{algorithm}
\SetAlgoLined
\caption{High level algorithm}

\textbf{Input:} F (randomly initialized)\;
augmented-lookalike-dataset = []\;
 \For{meta-epoch in meta-epochs}{
  $F$ = Alg-F($F$, augmented-lookalike-dataset)\;
  \For {iteration from 1..10000}{
    identity = choose-random-identity(CelebA)\;
    $i_{GT}$, $i'_{GT}$ = choose-random-images(identity)\;
    $i_{LR}$, $i'_{LR}$ = downscale $i_{GT}$, $i'_{GT}$\;
    z = optimize $\mathcal{L}_{SR} + \mathcal{L}_{sim}(i_{LR})$ and set $i = G(z)$\;
    z' = optimize $\mathcal{L}_{SR} + \mathcal{L}_{sim}(i_{LR}')$ and set $i = G(z')$\;
    augmented-lookalike-dataset.add(($i_{GT}$, $i'_{GT}$, $i$, $i'$))\;
  }
 }
 \label{alg:high_level}
\end{algorithm}
Specifically choose-random-identity chooses a random person from the CelebA dataset and choose-random-images chooses two random images with this person's identity. 
In our implementation we used 5 meta-epochs for Algorithm \ref{alg:high_level} and 100 epochs for Algorithm \ref{alg:alg_f}. 

\begin{table}
\begin{center}
 \begin{tabular}{ccc} 
 \toprule
$F$ & Loss & RE-ID Accuracy \\ \midrule
- & $\mathcal{L}_{SR}$ &70.5\% \\
$F_{baseline}$ & $\mathcal{L}_{SR} + \mathcal{L}_{sim}(I_{LR})$ &70.9\% \\
$F_{ours}$ & $\mathcal{L}_{SR} + \mathcal{L}_{sim}(I_{LR})$ &79.6\% \\
$F_{ours}$ & $\mathcal{L}_{SR} + \mathcal{L}_{att}(a)$ &96.0\% \\
\bottomrule
\end{tabular}
\end{center}
\vspace{-0.4cm}
\caption{Attribute re-identification accuracy on 5 attributes.}
\label{tab:attr_tab_reid}
\end{table}

\section{Histogram of Angles}
\label{sec:angles}

\begin{figure*}
  \centering
  \includegraphics[width=0.85\linewidth]{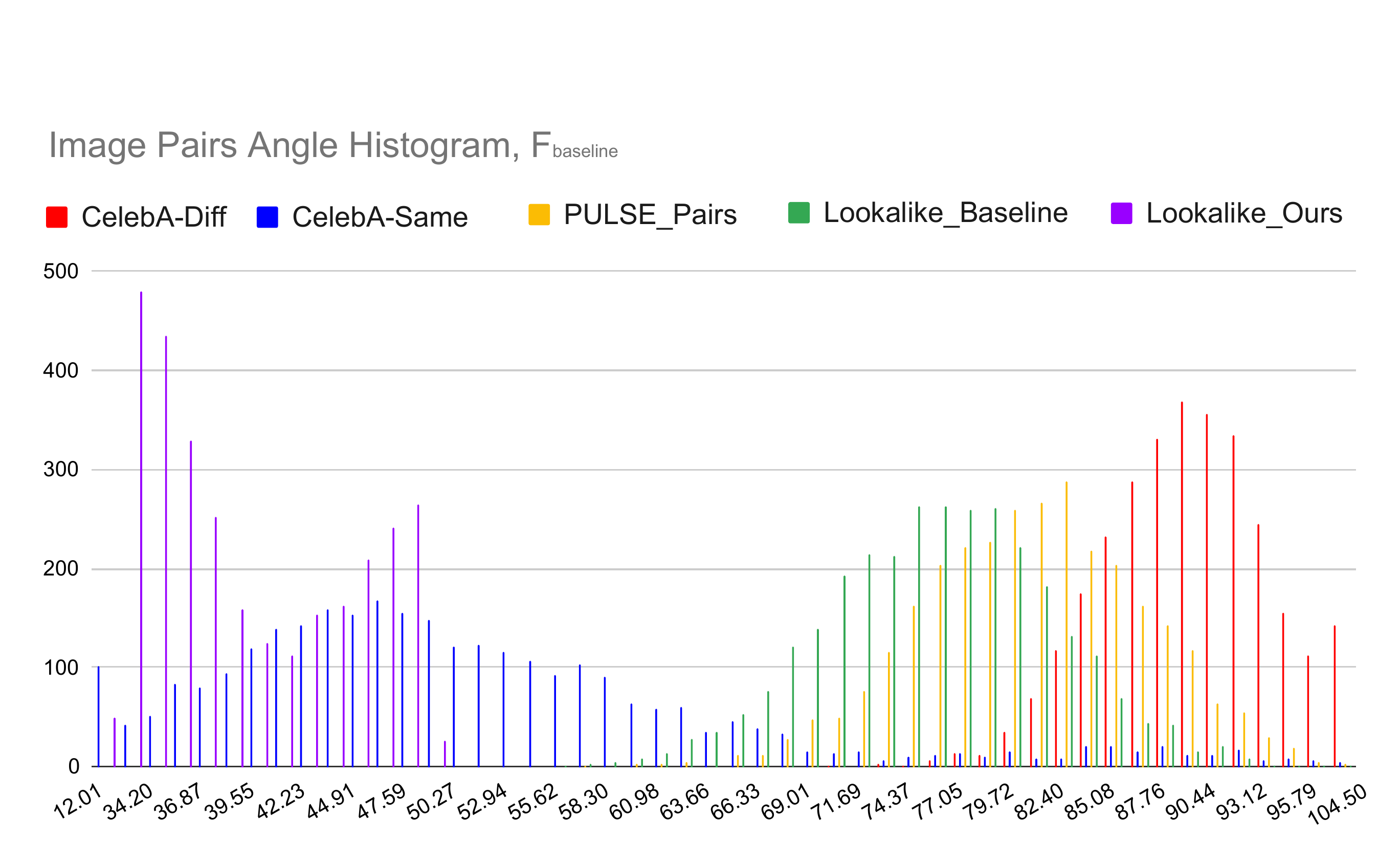} \\
  (a) \\
  \includegraphics[width=0.85\linewidth]{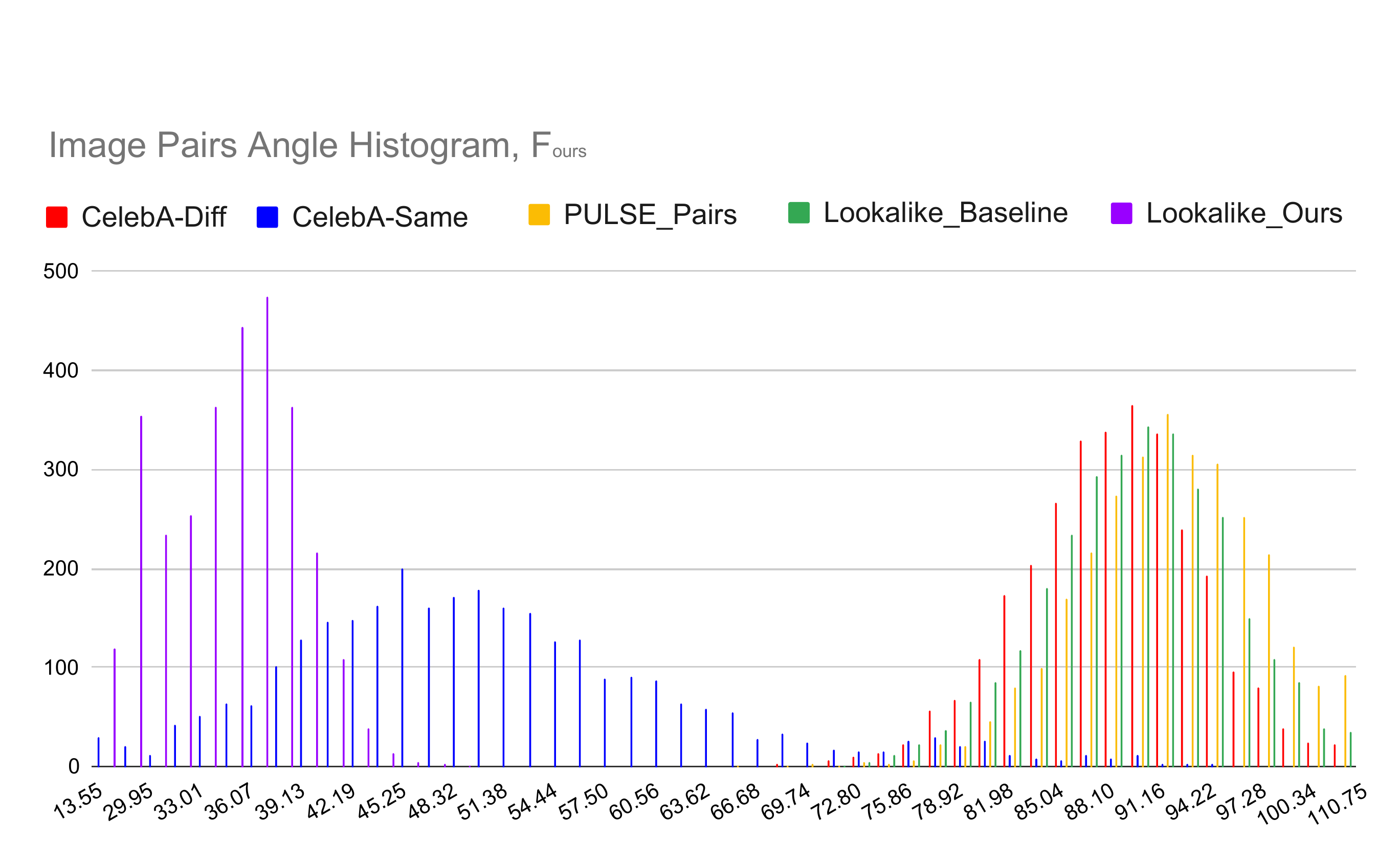} \\
  (b)
\caption{Comparison of angle histograms between the two face feature extractors. x-axis is the angle, and y axis is the frequency. In (a) $F_{baseline}$ is used and in (b) $F_{ours}$ is used. $F_{ours}$ generates a similar distribution for PULSE-Pairs, Lookalike-Baseline and CelebA-Diff datasets (green, yellow and red). 
This creates a successful separator at $\theta_{thr}=70^\circ$.}

\label{fig:histogram}
\end{figure*}

Tab.~1 of the main paper shows the mean and std of angles generated by $F_{baseline}$ and $F_{ours}$ on several different datasets. Fig.~\ref{fig:histogram} shows the histograms of angles. As can be seen $F_{ours}$ achieves superior separation of datasets.

\section{Attribute detection imbalance}
\label{sec:imbalance}

In Sec.~\ref{sec:extractor}, we mention the use of Inverse Class
Frequency as part of the attribute classifier $C_A$'s training. Without Inverse Class
Frequency, $34$ our of the $40$ attributes had an average precision (AP) below 60\%. Upon closer inspection this resulted in $C_A$ simply predicting the most common label of the attribute. 

Using Inverse Class
Frequency resulted in an increase of the mean average precision (mAP) from 54\% to 81\%, with none of the attributes falling below 60\% AP.

\end{document}